\documentclass[letterpaper]{article} % DO NOT CHANGE THIS
\usepackage{aaai2026}  % DO NOT CHANGE THIS
\usepackage{times}  % DO NOT CHANGE THIS
\usepackage{helvet}  % DO NOT CHANGE THIS
\usepackage{courier}  % DO NOT CHANGE THIS
\usepackage[hyphens]{url}  % DO NOT CHANGE THIS
\usepackage{graphicx} % DO NOT CHANGE THIS
\usepackage{natbib}  % DO NOT CHANGE THIS AND DO NOT ADD ANY OPTIONS TO IT
\usepackage{caption} % DO NOT CHANGE THIS AND DO NOT ADD ANY OPTIONS TO IT
\usepackage{color}
\usepackage{algorithm}
\usepackage{algorithmic}
\usepackage{booktabs}
\usepackage{amsmath}
\usepackage{amsfonts}

\usepackage{multirow}
\usepackage{newfloat}
\usepackage{listings}
\usepackage{newfloat}
\usepackage{listings}
\DeclareCaptionStyle{ruled}{labelfont=normalfont,labelsep=colon,strut=off} % DO NOT CHANGE THIS
\floatstyle{ruled}
\newfloat{listing}{tb}{lst}{}
\floatname{listing}{Listing}
\title{HOID-R1: Reinforcement Learning for Open-World Human-Object Interaction Detection Reasoning with Multimodal Large Language Model}
\author {
    Zhenhao Zhang\textsuperscript{\rm 1}\equalcontrib,
    Hanqing Wang\textsuperscript{\rm 2,3}\thanks{Project Leader.}\equalcontrib,
    Xiangyu Zeng\textsuperscript{\rm 3,4}\equalcontrib,\\
    Ziyu Cheng\textsuperscript{\rm 1,5},
    Jiaxin Liu\textsuperscript{\rm 1},
    Haoyu Yan\textsuperscript{\rm 1},
    Zhirui Liu\textsuperscript{\rm 1},
    Kaiyang Ji\textsuperscript{\rm 1},\\
    Tianxiang Gui\textsuperscript{\rm 1},
    Ke Hu\textsuperscript{\rm 1},
    Kangyi Chen\textsuperscript{\rm 1},
    Yahao Fan\textsuperscript{\rm 1},
    Mokai Pan\textsuperscript{\rm 1}
}
\affiliations {
    \textsuperscript{\rm 1}ShanghaiTech University,
    \textsuperscript{\rm 1}The Hong Kong University of Science and Technology (GZ),
    \textsuperscript{\rm 3}Shanghai AI Lab,\\
    \textsuperscript{\rm 4}Nanjing University
    \textsuperscript{\rm 5}University of Wisconsin, Madison\\
    zhangzhh2024@shanghaitech.edu.cn,hwang201@connect.hkust-gz.edu.cn
}

\usepackage{bibentry}
\begin{document}

\maketitle

\begin{abstract}
Understanding and recognizing human–object interaction (HOI) is a pivotal application in AR/VR and robotics. Recent open‑vocabulary HOI detection approaches depend exclusively on large language models for richer textual prompts, neglecting their inherent 3D spatial understanding capabilities. To address this shortcoming, we introduce \textbf{HOID-R1}, the first HOI detection framework that integrates chain-of-thought (CoT) guided supervised fine-tuning (SFT) with group relative policy optimization (GRPO) within a reinforcement learning (RL) paradigm. Specifically, we initially apply SFT to imbue the model with essential reasoning capabilities, forcing the model to articulate its thought process in the output. Subsequently, we integrate GRPO to leverage multi‑reward signals for policy optimization, thereby enhancing alignment across diverse modalities. To mitigate hallucinations in the CoT reasoning, we introduce an "MLLM‑as‑a‑judge" mechanism that supervises the CoT outputs, further improving generalization. Extensive experiments show that \textbf{HOID-R1} achieves state‑of‑the‑art performance on HOI detection benchmarks and outperforms existing methods in open‑world generalization to novel scenarios.

\end{abstract}

\section{Introduction}

Human–object interaction (HOI) detection seeks not only to localize human and object instances in visual scenes, but also to characterize the semantic and functional relationships that define their interactions. As a foundational component of human‐centric AI, accurate HOI detection underpins a diverse range of downstream applications—among them dexterous assistive and collaborative robotics, immersive augmented and virtual reality, surveillance and anomaly detection, advanced video understanding, and anticipatory activity forecasting. By modeling affordances, intentions, and social context, HOI detection endows autonomous agents with the perceptual and reasoning capabilities required for safe, effective operation in complex, human‐populated environments.

Existing HOI detection methods are predominantly confined to small-scale closed-set benchmarks (e.g., \cite{rw-2stage-ican}, \cite{rw-1stage-pointbase-ppdm}). In these benchmarks, models are trained and evaluated on a fixed set of interaction categories. This constraint yields limited out-of-distribution generalization. When faced with novel verbs, unseen objects, or previously unobserved interaction combinations, performance degrades sharply.

\begin{figure}
\centering 
\includegraphics[width=\linewidth]{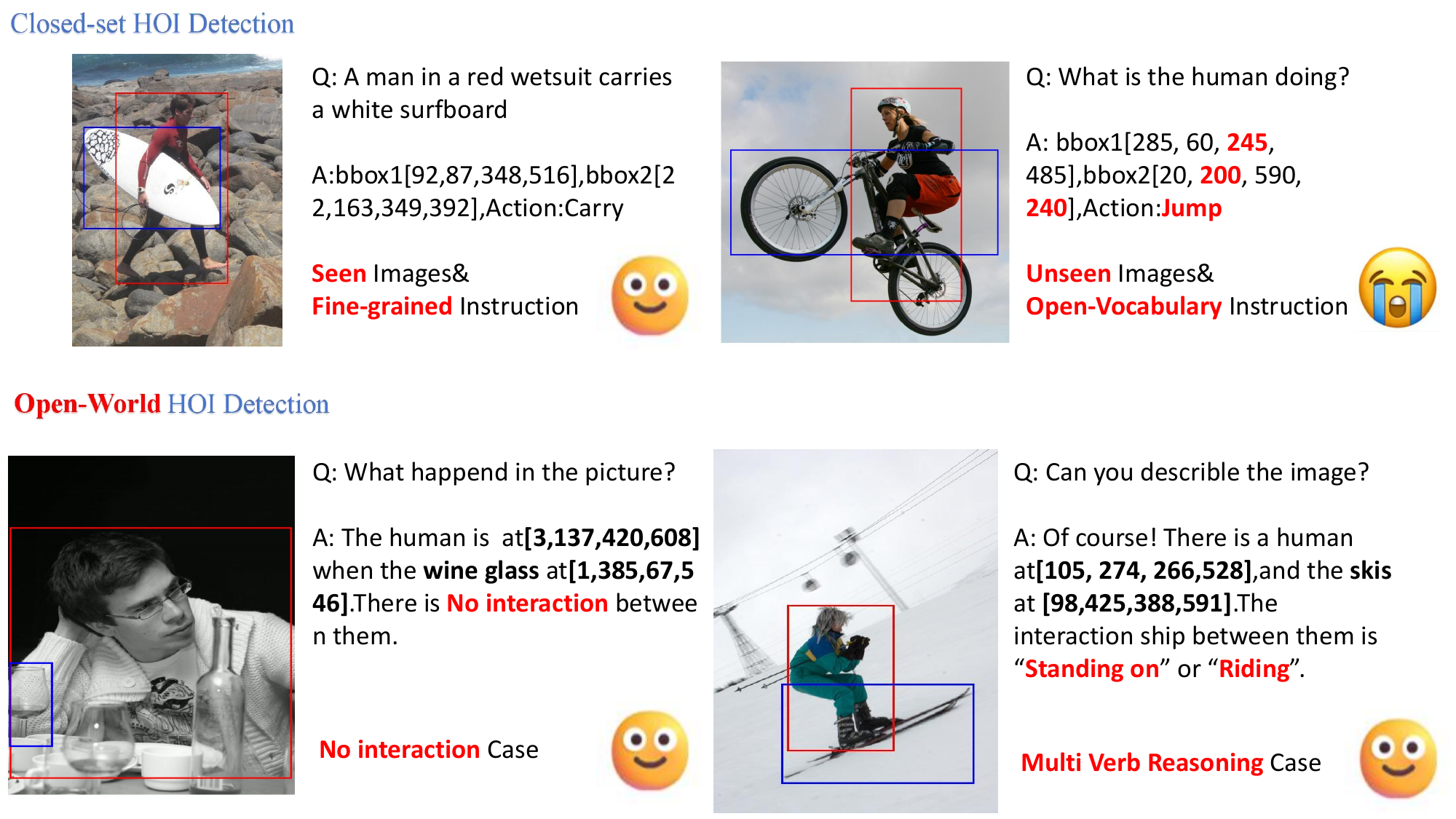}
\caption{\textbf{Motivation.} Closed‐set HOI detectors fail to generalize to novel verbs, objects, or interaction combinations in real scenes; adopting an open‐world paradigm enables structured reasoning and semantic supervision for robust zero‐shot inference under free‐form instructions.
}
\label{motivation} 
\end{figure}
Recent work has begun to exploit large vision language models to generate richer interaction prompts and enable zero-shot inference (e.g., \cite{rw-hoi-clip}). These prompt-based methods leverage only the linguistic priors of the underlying model and largely ignore its inherent reasoning capabilities\cite{dag}\cite{sdeval}. As a result, they remain sensitive to the precise phrasing of queries and struggle to disambiguate fine-grained or underspecified interactions. In contrast, our framework applies supervised fine-tuning followed by targeted post-training to fully harness the model’s reasoning power. The resulting open world HOI detector generalizes robustly across novel verbs, unseen objects, and fuzzy natural language queries.

To address these challenges, we propose \textbf{HOID-R1}, an open-world HOI detection framework that integrates multi-stage reasoning, visual grounding, and policy learning under continual semantic supervision. Given an input image and a free-form language query, our reasoning module produces a structured chain of thought\cite{Introduction-CoT-koj},\cite{introduction-CoT-jason} comprising sequential hypotheses about potential human-object interactions, while a parallel segmentation module localizes regions relevant to the task in the visual scene. These symbolic cues and pixel-level signals are then combined by a policy model trained with Generalized Reward Policy Optimization to generate spatial coordinates and interaction labels. During training, multiple reward functions assess physical plausibility, spatial consistency, and task accuracy, and a collection of vision language models acting as multimodal judges provides iterative feedback on intermediate reasoning steps. This supervision mechanism leverages the reasoning capacity of large-scale vision language models to identify and correct hallucinated or unsupported chains of thought trajectories, ensuring that every inference is grounded in both visual evidence and linguistic context. As a result, HOID-R1 achieves robust open vocabulary generalization and maintains high accuracy on novel verbs, unseen objects, and underspecified queries in real-world HOI scenarios.

In summary, our contributions are as follows:

\begin{itemize}
  \item \textbf{The first RL-based Chain-of-Thought framework for HOI detection.}  
    We introduce the first reinforcement learning paradigm that integrates a chain-of-thought reasoning process directly into HOI detection, enabling the model to decompose complex interaction queries into a sequence of sub-reasoning steps and learn to optimize each step via policy gradient.
  
  \item \textbf{Open-World HOI detection reasoning and multi-level dataset.}  
    Through supervised fine-tuning and GRPO-based post-training of the VLM, our model acquires Open-World human-object interaction reasoning capabilities and demonstrates strong generalization to open-vocabulary instructions and unseen images. To rigorously evaluate its open-world generalization, we hierarchically annotated a new HOI detection dataset

  \item \textbf{MLLM-as-a-Judge for chain-of-thought process.}  
    We leverage a pre-trained 3D multimodal large language model as a soft “judge” to evaluate and guide each reasoning step, preventing the model from arriving at correct conclusions through flawed reasoning processes. We have enhanced its reasoning capabilities.

  \item \textbf{State-of-the-art performance.}  Extensive experiments on HICO-DET and SWIG-HOI show that our approach outperforms existing baselines by a significant margin across all key metrics, achieving new state-of-the-art results in both seen- and unseen-object HOI detection.

\end{itemize}

\section{Related Work}

\subsection{HOI Detection} Current HOI detection methods can be mainly divided into two categories: two-stage paradigm\cite{rw-2stage-ican},\cite{rw-hoi-2stage-ex} and one-stage paradigm\cite{rw-1stage-anchor-uniondet},\cite{rw-1stage-pointbase-ppdm},\cite{rw-1stage-set},\cite{rw-1stage-set-Qpic}. Two-stage strategy first detects all human and object instances using a pre-trained object detector, such as Faster R-CNN\cite{rw-2stage-ican}, followed by a second-stage module that enumerates possible human-object pairs and predicts their interactions. Although this design benefits from the maturity and robustness of standalone object detectors, it suffers from the inefficiency of exhaustive pairwise matching and limited ability to model contextual dependencies among entities. In contrast, one-stage approach detects (human-object-interaction) triplets directly through different perspectives, e.g., point-based detection\cite{rw-1stage-pointbase-ppdm} formulates HOI triplets as pairs of keypoints, such as the center points of human and object bounding boxes, and each interaction is modeled by predicting a pair of spatial points along with verb classification, anchor-based detection\cite{rw-1stage-anchor-uniondet} extends the concept of anchor boxes from object detection to interaction modeling, human and object entities are predicted based on predefined anchor regions, and interactions are inferred using features extracted from the union area of the predicted boxes and set prediction methods\cite{rw-1stage-set},\cite{rw-1stage-set-Qpic} reformulate the HOI detection problem as a set-to-set matching problem, thus avoiding human-object pairing.
\begin{figure*}[t]
    \centering
    \includegraphics[width=\linewidth]{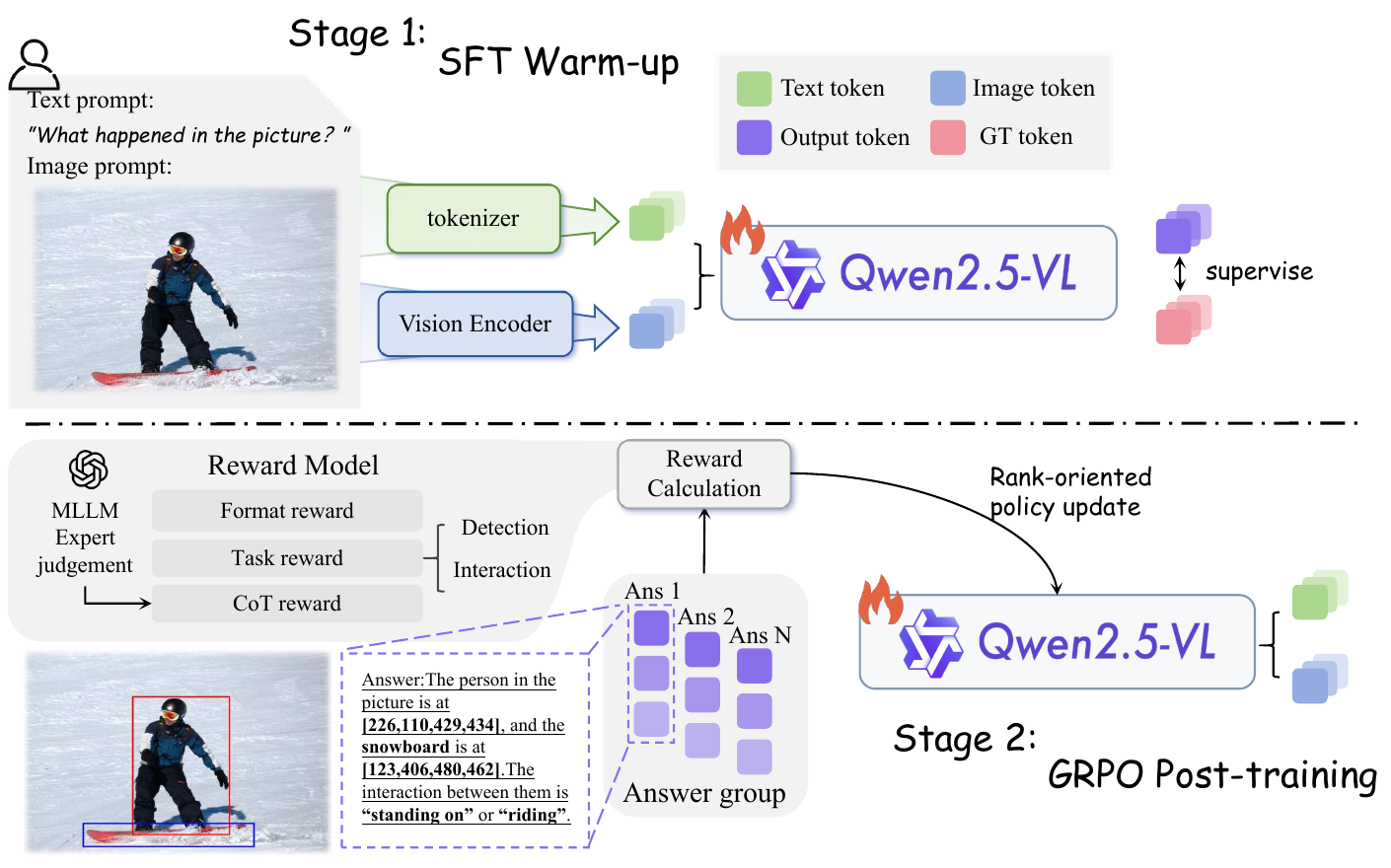}
    \caption{\textbf{Pipeline.} HOID-R1 first encodes the input image and free-form query into multimodal embeddings and warms up via SFT to generate chain-of-thought–annotated HOI triplets. During GRPO post-training, it samples candidate triplets, uses an MLLM judge to compute a composite reward on format compliance, detection accuracy, interaction classification, and CoT coherence, and updates the policy for precise localization and faithful reasoning.
}
\end{figure*}
\subsection{Large Reasoning Model}  Unlike traditional LLMs, which can only take textual inputs, Multimodal Large Language Models (MLLMs) extend the capabilities of traditional large language models by integrating information from multiple modalities, such as text, images, audio, and video. By jointly modeling cross-modal interactions, MLLMs\cite{affr1} enable a wide range of tasks, including visual question answering, image captioning, and multimodal reasoning. The emergence of Large Reasoning Models (LRMs)\cite{rw-r1},\cite{rw-r1-grpo},\cite{rw-r1-hoi},\cite{rw-r1-motion},\cite{rw-r1-vision}, which are explicitly designed to enhance reasoning capabilities beyond language generation. One prominent example is Deepseek R1, a reasoning-centric model that integrates both pretraining on reasoning-oriented data and instruction tuning to excel at tasks requiring systematic thought. Compared to traditional LLMs, LRMs\cite{spbme} like R1 are better at decomposing complex problems, following long-term logical chains, and aligning intermediate steps with final outputs. LRMs have now being used in many topics: in motion generation tasks\cite{rw-r1-motion}, LRMs can reasoning over temporal sequences and physical constraints, thus synthesize realistic and controllable human motion trajectories from abstract instructions or sparse keyframes, in computer vision\cite{rw-r1-vision}, VLM-R1 demonstrates strong generalization across diverse tasks such as visual question answering, image captioning, and referring expression comprehension and also by modeling multi-entity interactions through structured reasoning, LRMs can better capture the semantic dependencies between humans, objects, and actions in HOI detection tasks\cite{rw-r1-hoi}. 

\section{Method}
We propose a unified framework that integrates three components. An HOI detection network learns to localize interactions and classify actions and objects using intersection‐over‐union and classification rewards. A chain of thought generator produces intermediate reasoning steps that are scored by a Process Reward Model and a Generalizable Reward Model to form a reasoning reward. A pretrained multimodal LLM judge uses these two models to provide mixed supervision at both individual steps and groups of steps. All rewards are combined into a single objective and optimized via reinforcement learning, yielding precise localization, accurate classification, and coherent reasoning.

\subsection{Supervised Fine-Tuning}
To bridge the gap between general reasoning capabilities and the structured requirements of HOI detection outputs, we employ Supervised Fine-Tuning (SFT) as an essential preparatory stage. Directly applying general-purpose reasoning models such as DeepSeek-R1 to HOI-related tasks often leads to inconsistently formatted outputs, especially when generating scene graph-style captions or structured interaction tuples. To address this, we design an SFT stage where the model is guided to adhere strictly to the desired output format (e.g., ⟨subject, verb, object⟩ triplets\cite{methods-verb-object-tri}), using curated demonstrations.

However, relying solely on rigid format supervision may inadvertently suppress the model’s intrinsic reasoning ability, reducing it to pattern-matching instead of true inference. To counteract this, we introduce Cognitive Chain-of-Thought (CoT)\cite{methods-relation-r1} prompting within the SFT stage. Here, stepwise reasoning processes grounding them in the visual input, and inferring their interaction, are explicitly annotated using special <think> tags. This encourages the model to internalize a cognitively grounded reasoning procedure while still learning to output syntactically and semantically structured HOI predictions. Empirically, this hybrid supervision strategy enhances both interpretability and output consistency.

We impose task-specific constraints to reflect the domain characteristics of HOI detection, we specifies the domain of the thinking process: 1) human detection(the first entity of the CoT is human); 2) object detection(the second entity is restrict to objects); 3) relation existence(we tried to limit the relation within several verbs). This guides the model to focus its reasoning within a valid interaction space, reducing hallucination and improving output faithfulness. Furthermore, this constraint-aware design helps prevent overfitting to idiosyncratic CoT patterns, ensuring better generalization across diverse scenes.

Empirically, this hybrid strategy—combining structured format supervision with constrained yet expressive CoT reasoning—significantly improves both the interpretability and consistency of HOI predictions.

\subsection{MLLM-as-a-Judge}
Large language models often produce correct final answers while their intermediate reasoning contains semantic deviations or logical flaws. To address this problem, we introduce MLLM-as-a-Judge\cite{methods-mllm-judge}, which uses a pretrained multimodal large language model to provide mixed supervision over the chain of thought generation. 

\textbf{Process Reward Model (PRM):}The PRM assigns a score or a binary judgment (“Is this step correct?”) at each reasoning step in the generated CoT, and uses that as the training signal to fine-tune the model. In other words, the PRM provides correctness/incorrectness labels or scores for every intermediate step, teaching the model to be reliable at each stage of its reasoning.

\textbf{Generalizable Reward Model (GRM):}Building on standard preference learning, the GRM additionally preserves and regularizes the reward model’s generative ability—applying a language-modeling loss to its hidden states—so that it can both evaluate the quality of reasoning and generate coherent intermediate steps like a language model. This generative supervision significantly boosts the reward model’s generalization to unseen tasks or out-of-distribution samples.

During training, the student model’s generated reasoning chain is fed into the MLLM judge, which generates step-level and generalizable feedback signals through PRM and GRM, and combines these signals into a composite reward within a reinforcement learning framework. By enforcing constraints on the intermediate reasoning process, the MLLM-as-a-Judge mechanism effectively suppresses tendencies to arrive at correct outcomes through flawed logic, thereby improving the reliability and interpretability of the model’s reasoning.
\subsection{Group Relative Policy Optimization}
To optimize the R1 framework, the group relative policy optimization is adopted. GRPO is a reinforcement learning (RL) algorithm designed for optimizing policy models.  As proposed in DeepSeek-R1\cite{rw-r1}, GRPO eliminates dependency on critic networks by leveraging direct response comparisons within stochastically sampled output groups. This approach reduces computational overhead while maintaining optimization stability. The objective function of GRPO  leverages direct pairwise comparisons within the sampled group through implicit reward normalization. By using the group mean as a dynamic baseline, it reduces gradient variance while maintaining optimization stability. Crucially, it eliminates per-token value estimation, cutting computational overhead by $\mathcal{O}(n)$ for sequence length $n$ compared to critic-dependent methods:

\begin{equation}
A = \frac{1}{G} \sum_{i=1}^G\min\left(\rho_iA_i,\mathrm{clip}(\rho_i,1-\epsilon,1+\epsilon)A_i\right)
\end{equation}

\begin{equation}
B=\beta D_{\mathrm{KL}}(\pi_\theta\|\pi_{\mathrm{ref}})
\end{equation}

\begin{equation}
J_{\mathrm{GRPO}}(\theta) =  \mathbf{E}_{q\sim\mathcal{Q},\{o_i\}_{i=1}^G\sim\pi_{\theta_{\mathrm{old}}}}\left[A-B\right]
\end{equation}

where $\rho_i=\frac{\pi_\theta(o_i|q)}{\pi_{\theta_{old}}(o_i|q)}$ quantifies policy shift between current policy $\pi_\theta$ and behavioral policy $\pi_{\theta_{\mathrm{old}}}$., $\epsilon$ controls clipping thresholds and $\beta$ indicates the deviations via the KL divergence term. The advantage score $A_i$ mitigates reward scale sensitivity and reduces gradient variance, which can standardize rewards to stabilize training:

\begin{equation}
A_i=\frac{r_i-\text{mean}(\{r_1,...,r_c\})}{\text{std}(\{r_1,...,r_c\})}
\end{equation}

$r_i$ represents the reward for response $o_i$. The KL Divergence is defined to control exploration without too much divergence from $\pi_{ref}$ as follows:

\begin{equation}
D_{KL}(\pi_\theta||\pi_{ref})=\frac{\pi_{ref}(o_i|q)}{\pi_\theta(o_i|q)}-\log(\frac{\pi_{ref}(o_i|q)}{\pi_\theta(o_i|q)}) - 1
\end{equation}

Our reward design generally consists of four parts: Format Reward, Detection Reward, Interaction Reward, and CoT Reward.

\textbf{Format Reward} This reward is designed to restrict the reasoning template format:  Thus, the reward design is:

\begin{equation}
r_{format}(o_i)=
\begin{cases}
1 & \text{if } o_i \text{format is right} \\
0 & \text{otherwise}
\end{cases}
\end{equation}
This binary function thus enforces format requirements on syntax while granting flexibility in content.

\textbf{Detection Reward} Inspired by established object detection practices, we devise the sample‐level IoU score $R_{\mathrm{IoU}}$ and the sample‐level regression accuracy $R_{\mathrm{reg}}$ in the predicted anchor boxes for each sample\cite{methods-detection-reward},\cite{methods-detection-reward-Fast-cnn}. The former measures the fraction of anchors whose Intersection over Union with their ground‐truth boxes reaches or exceeds 0.5; the latter measures the fraction of anchors whose normalized L1 coordinate error falls below a threshold $\delta$. Specifically, an anchor prediction is deemed correct if:

\begin{itemize}
  \item \textbf{Overlap Accuracy} The predicted anchor box achieves an Intersection over Union of at least 0.5 with its ground‐truth counterpart.
  \item \textbf{Coordinate Precision} The normalized L1 distance between predicted and ground‐truth box coordinates is below the threshold $\delta$.
\end{itemize}

The final detection reward is computed as a weighted combination of these two metrics:
\begin{equation}
r_{\mathrm{det}}(o) = \beta \cdot R_{\mathrm{IoU}} + (1 - \beta)\cdot R_{\mathrm{reg}},
\end{equation}
where $\beta$ balances the trade‐off between overlap accuracy and coordinate precision.
\begin{table*}[!htbp]
\centering
\begin{tabular}{@{}lcccccccc@{}}
\toprule
\textbf{Model} & \multicolumn{4}{c}{\textbf{Seen}} & \multicolumn{4}{c}{\textbf{Unseen}} \\
\cmidrule(lr){2-5} \cmidrule(lr){6-9}
 & \textbf{H-mIOU$\uparrow$} & \textbf{O-mIOU$\uparrow$}  & \textbf{A-ACC$\uparrow$}& \textbf{mAP$\uparrow$} & \textbf{H-mIOU$\uparrow$} & \textbf{O-mIOU$\uparrow$}  & \textbf{A-ACC$\uparrow$}& \textbf{mAP$\uparrow$} \\
\midrule

\multicolumn{9}{@{}l}{\textit{\textbf{Fine-grained annotation}}} \\
HOI-Trans &0.56$^{\pm0.018}$  &0.54$^{\pm0.015}$  &0.60$^{\pm0.028}$  &33.64$^{\pm1.27}$ &0.51$^{\pm0.021}$  &0.52$^{\pm0.019}$  &0.56$^{\pm0.025}$  &31.28$^{\pm1.34}$ \\
DiffHOI &0.63$^{\pm0.023}$  &0.61$^{\pm0.020}$  &0.67$^{\pm0.027}$ &44.06$^{\pm1.91}$  &0.61$^{\pm0.024}$  &0.60$^{\pm0.018}$  &0.63$^{\pm0.026}$ &42.25$^{\pm1.68}$ \\
PAFR &0.65$^{\pm0.025}$  &0.66$^{\pm0.022}$  &0.72$^{\pm0.029}$ &51.22$^{\pm2.39}$  &0.63$^{\pm0.026}$  &0.62$^{\pm0.020}$  &0.67$^{\pm0.031}$ &47.51$^{\pm1.89}$ \\
HORP &0.67$^{\pm0.021}$  &0.69$^{\pm0.024}$  &0.70$^{\pm0.032}$ &52.75$^{\pm1.84}$  &0.66$^{\pm0.027}$  &0.66$^{\pm0.023}$  &0.69$^{\pm0.030}$ &49.03$^{\pm2.05}$ \\
\textbf{Ours} & \textbf{0.72$^{\pm0.026}$}  & \textbf{0.73$^{\pm0.028}$} & \textbf{0.79$^{\pm0.031}$} & \textbf{55.07$^{\pm2.11}$}  & \textbf{0.71$^{\pm0.025}$} & \textbf{0.70$^{\pm0.027}$} & \textbf{0.75$^{\pm0.030}$} & \textbf{52.98$^{\pm2.18}$} \\
\midrule

\multicolumn{9}{@{}l}{\textit{\textbf{Precise annotation}}} \\
HOI-Trans &0.55$^{\pm0.019}$ &0.54$^{\pm0.021}$ &0.58$^{\pm0.028}$ &32.46$^{\pm1.12}$ &0.50$^{\pm0.022}$ &0.51$^{\pm0.017}$ &0.54$^{\pm0.024}$ &29.55$^{\pm1.26}$ \\
DiffHOI &0.62$^{\pm0.020}$ &0.61$^{\pm0.025}$ &0.65$^{\pm0.031}$ &42.17$^{\pm1.53}$ &0.60$^{\pm0.019}$ &0.59$^{\pm0.024}$ &0.61$^{\pm0.027}$ &40.05$^{\pm1.78}$ \\
PAFR &0.64$^{\pm0.022}$ &0.65$^{\pm0.026}$ &0.70$^{\pm0.034}$ &49.68$^{\pm2.27}$ &0.62$^{\pm0.021}$ &0.61$^{\pm0.022}$ &0.65$^{\pm0.030}$ &46.33$^{\pm2.05}$ \\
HORP &0.66$^{\pm0.018}$ &0.68$^{\pm0.020}$ &0.69$^{\pm0.028}$ &51.46$^{\pm1.76}$ &0.64$^{\pm0.023}$ &0.65$^{\pm0.021}$ &0.67$^{\pm0.031}$ &47.65$^{\pm1.99}$ \\
\textbf{Ours} & \textbf{0.70$^{\pm0.025}$} & \textbf{0.72$^{\pm0.027}$} & \textbf{0.76$^{\pm0.030}$} & \textbf{53.81$^{\pm2.13}$} & \textbf{0.71$^{\pm0.026}$} & \textbf{0.69$^{\pm0.025}$} & \textbf{0.72$^{\pm0.029}$} & \textbf{51.19$^{\pm2.21}$} \\
\midrule

\multicolumn{9}{@{}l}{\textit{\textbf{Open-Vocabulary annotation}}} \\
HOI-Trans &0.54$^{\pm0.020}$ &0.52$^{\pm0.018}$ &0.57$^{\pm0.025}$ &29.17$^{\pm1.13}$ &0.49$^{\pm0.022}$ &0.50$^{\pm0.019}$ &0.52$^{\pm0.023}$ &28.90$^{\pm1.07}$ \\
DiffHOI &0.61$^{\pm0.023}$ &0.60$^{\pm0.024}$ &0.64$^{\pm0.030}$ &40.95$^{\pm1.55}$ &0.59$^{\pm0.025}$ &0.58$^{\pm0.021}$ &0.60$^{\pm0.026}$ &39.44$^{\pm1.65}$ \\
PAFR &0.63$^{\pm0.025}$ &0.63$^{\pm0.022}$ &0.68$^{\pm0.031}$ &45.28$^{\pm2.06}$ &0.61$^{\pm0.021}$ &0.60$^{\pm0.020}$ &0.63$^{\pm0.030}$ &43.78$^{\pm1.74}$ \\
HORP &0.65$^{\pm0.020}$ &0.66$^{\pm0.025}$ &0.67$^{\pm0.029}$ &46.09$^{\pm2.12}$ &0.63$^{\pm0.024}$ &0.63$^{\pm0.021}$ &0.65$^{\pm0.028}$ &45.12$^{\pm2.04}$ \\
\textbf{Ours} & \textbf{0.69$^{\pm0.024}$} & \textbf{0.71$^{\pm0.027}$} & \textbf{0.75$^{\pm0.032}$} & \textbf{51.68$^{\pm2.36}$} & \textbf{0.68$^{\pm0.023}$} & \textbf{0.68$^{\pm0.022}$} & \textbf{0.72$^{\pm0.030}$} & \textbf{50.03$^{\pm2.14}$} \\
\bottomrule
\end{tabular}
\caption{
\textbf{Main Result on HICO-DET.}
}
\label{main-hico}
\end{table*}

\textbf{Interaction Reward:}Inspired by established classification practices, we define the sample-level action accuracy $R_{\mathrm{act}}$ and the sample-level object accuracy $R_{\mathrm{obj}}$ for each sample. The first metric measures the fraction of samples whose predicted action label matches the ground truth action. The second metric measures the fraction of samples whose predicted object label matches the ground truth object. Specifically, an interaction prediction is correct if:

\begin{itemize}
  \item \textbf{Action Correctness}: The model’s predicted action label (for example “pick up”, “pour”, or “rotate”) must match the ground truth action label exactly. A sample is counted as an action correct only when the predicted category corresponds to the annotated category.
  \item \textbf{Object Correctness}: The model’s predicted object label (for example “cup”, “bottle”, or “book”) must match the ground truth object label exactly. A sample is considered object correct only when the predicted name aligns with the annotated object name unambiguously.
\end{itemize}

The final interaction reward is computed as a weighted combination of these two metrics:
\begin{equation}
r_{\mathrm{int}}(o) = \gamma \cdot R_{\mathrm{act}} + (1 - \gamma) \cdot R_{\mathrm{obj}},
\end{equation}
where $\gamma$ balances the importance of action correctness and object correctness.

\textbf{CoT Reward} To encourage both fine-grained relevance and high-level coherence in the model’s intermediate reasoning\cite{methods-cot-reward}, we design a \emph{Chain-of-Thought} reward $r_{\mathrm{CoT}}$ that integrates two complementary signals:

\begin{itemize}
  \item \textbf{Process Reward Model (PRM).}  \cite{methods-prm}
    Let $N$ be the number of steps in the generated CoT, and let $s_i \in [0,1]$ be the PRM score for step $i$. We define the step‐level reward
    \begin{equation}
      R_{\mathrm{prm}}
      = \frac{1}{N} \sum_{i=1}^{N} s_i.
    \end{equation}
  \item \textbf{Generalizable Reward Model (GRM).}  
    Partition the chain into $M$ groups of consecutive steps, and let $g_j \in [0,1]$ be the GRM score for group $j$\cite{methods-grm-1},\cite{methods-grm-2}. We define the group‐level reward
    \begin{equation}
      R_{\mathrm{grm}}
      = \frac{1}{M} \sum_{j=1}^{M} g_j.
    \end{equation}
\end{itemize}

These two signals are combined into a single scalar reward:
\begin{equation}
  r_{\mathrm{CoT}}
  = \lambda\,R_{\mathrm{prm}} \;+\;(1 - \lambda)\,R_{\mathrm{grm}},
\end{equation}
where $\lambda \in [0,1]$ balances the emphasis between step‐level accuracy and group‐level coherence.

By optimizing this reward within a reinforcement learning framework, the model is encouraged to produce reasoning trajectories that are both semantically aligned with the prompt and logically coherent throughout.

\begin{table*}[!htbp]
\centering
\begin{tabular}{@{}lcccccccc@{}}
\toprule
\textbf{Model} & \multicolumn{4}{c}{\textbf{Seen}} & \multicolumn{4}{c}{\textbf{Unseen}} \\
\cmidrule(lr){2-5} \cmidrule(lr){6-9}
 & \textbf{H-mIOU$\uparrow$} & \textbf{O-mIOU$\uparrow$} & \textbf{A-ACC$\uparrow$} & \textbf{mAP$\uparrow$} & \textbf{H-mIOU$\uparrow$} & \textbf{O-mIOU$\uparrow$} & \textbf{A-ACC$\uparrow$} & \textbf{mAP$\uparrow$} \\
\midrule

\multicolumn{9}{@{}l}{\textit{\textbf{Fine-grained annotation}}} \\
HOI-Trans &0.38$^{\pm0.012}$ &0.36$^{\pm0.014}$ &0.41$^{\pm0.019}$ &22.94$^{\pm1.08}$ &0.33$^{\pm0.016}$ &0.34$^{\pm0.015}$ &0.38$^{\pm0.016}$ &21.39$^{\pm1.06}$ \\
DiffHOI &0.41$^{\pm0.015}$ &0.39$^{\pm0.019}$ &0.43$^{\pm0.017}$ &29.93$^{\pm1.47}$ &0.40$^{\pm0.015}$ &0.39$^{\pm0.014}$ &0.41$^{\pm0.015}$ &28.47$^{\pm1.29}$ \\
PAFR &0.45$^{\pm0.016}$ &0.45$^{\pm0.014}$ &0.49$^{\pm0.021}$ &34.34$^{\pm1.50}$ &0.42$^{\pm0.015}$ &0.41$^{\pm0.019}$ &0.44$^{\pm0.018}$ &32.39$^{\pm1.42}$ \\
HORP &0.46$^{\pm0.018}$ &0.47$^{\pm0.015}$ &0.48$^{\pm0.015}$ &35.99$^{\pm1.71}$ &0.44$^{\pm0.017}$ &0.44$^{\pm0.014}$ &0.46$^{\pm0.018}$ &34.28$^{\pm1.58}$ \\
\textbf{Ours} &\textbf{0.49$^{\pm0.018}$} &\textbf{0.49$^{\pm0.019}$} &\textbf{0.55$^{\pm0.021}$} &\textbf{37.74$^{\pm1.88}$} &\textbf{0.47$^{\pm0.017}$} &\textbf{0.46$^{\pm0.017}$} &\textbf{0.49$^{\pm0.021}$} &\textbf{35.51$^{\pm1.73}$} \\
\midrule

\multicolumn{9}{@{}l}{\textit{\textbf{Precise annotation}}} \\
HOI-Trans &0.37$^{\pm0.012}$ &0.36$^{\pm0.013}$ &0.39$^{\pm0.018}$ &21.75$^{\pm1.07}$ &0.33$^{\pm0.015}$ &0.34$^{\pm0.015}$ &0.36$^{\pm0.017}$ &19.81$^{\pm0.96}$ \\
DiffHOI &0.40$^{\pm0.017}$ &0.39$^{\pm0.014}$ &0.42$^{\pm0.016}$ &28.02$^{\pm1.39}$ &0.39$^{\pm0.015}$ &0.38$^{\pm0.015}$ &0.40$^{\pm0.019}$ &26.23$^{\pm1.23}$ \\
PAFR &0.43$^{\pm0.016}$ &0.44$^{\pm0.016}$ &0.46$^{\pm0.020}$ &33.28$^{\pm1.66}$ &0.41$^{\pm0.017}$ &0.40$^{\pm0.015}$ &0.42$^{\pm0.021}$ &30.93$^{\pm1.44}$ \\
HORP &0.44$^{\pm0.019}$ &0.46$^{\pm0.017}$ &0.46$^{\pm0.019}$ &34.47$^{\pm1.59}$ &0.42$^{\pm0.017}$ &0.44$^{\pm0.017}$ &0.44$^{\pm0.020}$ &31.77$^{\pm1.43}$ \\
\textbf{Ours} &\textbf{0.47$^{\pm0.017}$} &\textbf{0.48$^{\pm0.018}$} &\textbf{0.50$^{\pm0.021}$} &\textbf{36.05$^{\pm1.81}$} &\textbf{0.47$^{\pm0.017}$} &\textbf{0.45$^{\pm0.016}$} &\textbf{0.48$^{\pm0.021}$} &\textbf{34.19$^{\pm1.54}$} \\
\midrule

\multicolumn{9}{@{}l}{\textit{\textbf{Open-Vocabulary annotation}}} \\
HOI-Trans &0.35$^{\pm0.015}$ &0.34$^{\pm0.014}$ &0.37$^{\pm0.017}$ &19.73$^{\pm0.99}$ &0.32$^{\pm0.015}$ &0.32$^{\pm0.013}$ &0.34$^{\pm0.017}$ &18.79$^{\pm0.92}$ \\
DiffHOI &0.40$^{\pm0.015}$ &0.39$^{\pm0.016}$ &0.42$^{\pm0.018}$ &27.02$^{\pm1.38}$ &0.38$^{\pm0.017}$ &0.37$^{\pm0.016}$ &0.39$^{\pm0.019}$ &25.63$^{\pm1.30}$ \\
PAFR &0.41$^{\pm0.017}$ &0.41$^{\pm0.017}$ &0.44$^{\pm0.019}$ &29.43$^{\pm1.51}$ &0.40$^{\pm0.016}$ &0.39$^{\pm0.016}$ &0.41$^{\pm0.017}$ &28.02$^{\pm1.38}$ \\
HORP &0.42$^{\pm0.017}$ &0.43$^{\pm0.016}$ &0.45$^{\pm0.019}$ &30.17$^{\pm1.61}$ &0.41$^{\pm0.017}$ &0.41$^{\pm0.017}$ &0.42$^{\pm0.018}$ &29.33$^{\pm1.47}$ \\
\textbf{Ours} &\textbf{0.45$^{\pm0.018}$} &\textbf{0.47$^{\pm0.017}$} &\textbf{0.49$^{\pm0.020}$} &\textbf{33.64$^{\pm1.72}$} &\textbf{0.44$^{\pm0.017}$} &\textbf{0.44$^{\pm0.017}$} &\textbf{0.46$^{\pm0.020}$} &\textbf{32.52$^{\pm1.59}$} \\
\bottomrule
\end{tabular}

\caption{
\textbf{Main Result on SWIG-HOI.}
}
\label{main-swig}
\end{table*}

\section{Experiments}

Our model is trained in two phases: (1) supervised fine-tuning on human–object interaction data; and (2) post-training using GRPO. We employ diverse datasets and evaluation metrics to assess our approach, demonstrating its capability to detect HOI under high-level instructions in both seen and unseen scenarios. Comparative experiments and ablation studies validate our design choices. All experiments are conducted on eight NVIDIA A800 GPUs.

\subsection{Dataset}
In our experiments, we selected two of the most commonly used benchmark datasets in the human-object interaction (HOI) detection field and re-annotated them.

\textbf{HICO-DET\cite{hico}} HICO-DET includes 47,776 images (38,118 for training, 9,658 for testing) annotated with 600 HOI categories formed by 117 verb classes and 80 object classes, totaling over 150,000 human–object pairs. Following standard zero-shot protocols, 120 of the rarest interaction triplets are withheld during training to assess a model’s ability to recognize unseen images 

\textbf{SWIG-HOI\cite{swig}} Assembled from the SWIG and DOH datasets, SWIG-HOI comprises roughly 45,000 training images and 14,000 test images, covering 406 human actions and 1,000 object categories. Its test split contains about 5,500 human–object interaction instances, of which nearly 1,800 interactions are not seen during training, making it a challenging benchmark for open-vocabulary HOI detection

We annotate two datasets using three distinct schemes: fine-grained annotation, precise annotation, and open‐vocabulary annotation. Each successive scheme places increasingly greater demands on the model’s generalization capabilities.

\begin{itemize}
    \item \textbf{Fine-grained annotation}: accurately describing the diverse attributes and actions of the people and objects depicted in the image. (e.g., "A man wearing black clothes is drinking a blue cup of water")
    \item \textbf{Precise annotation}: describing only the interactive actions and object depicted in the image. (e.g., "A man is drinking water")
    \item \textbf{Open‐vocabulary annotation}: providing only an open‐ended description of the person or object in the image (e.g., “What is the man doing?”, “What action is the cup performing?”), or posing a broad query (e.g., “What is happening in the image?”).
\end{itemize}
In addition, we partition the images into seen and unseen subsets and evaluate our method on each. The results show that our model achieves state-of-the-art performance across all six experimental settings defined by the three annotation schemes. Notably, it demonstrates strong generalization under open-vocabulary annotation on unseen images.

\begin{table*}[!htbp]
\centering
\begin{tabular}{@{}lcccccccc@{}}
\toprule
\textbf{Model} & \multicolumn{4}{c}{\textbf{Seen}} & \multicolumn{4}{c}{\textbf{Unseen}} \\
\cmidrule(lr){2-5} \cmidrule(lr){6-9}
 & \textbf{H-mIOU$\uparrow$} & \textbf{O-mIOU$\uparrow$} & \textbf{A-ACC$\uparrow$} & \textbf{mAP$\uparrow$} & \textbf{H-mIOU$\uparrow$} & \textbf{O-mIOU$\uparrow$} & \textbf{A-ACC$\uparrow$} & \textbf{mAP$\uparrow$} \\
\midrule

\multicolumn{9}{@{}l}{\textit{\textbf{Open-Vocabulary annotation}}} \\
W/O PT       & 0.54$^{\pm0.017}$ & 0.53$^{\pm0.018}$ & 0.57$^{\pm0.021}$ & 39.34$^{\pm1.92}$ & 0.54$^{\pm0.016}$ & 0.53$^{\pm0.019}$ & 0.57$^{\pm0.023}$ & 37.74$^{\pm1.75}$ \\
W/O FR      & 0.59$^{\pm0.020}$ & 0.62$^{\pm0.025}$ & 0.67$^{\pm0.031}$ & 43.94$^{\pm1.85}$ & 0.61$^{\pm0.019}$ & 0.60$^{\pm0.020}$ & 0.62$^{\pm0.029}$ & 42.91$^{\pm1.68}$ \\
W/O DR    & 0.60$^{\pm0.022}$ & 0.60$^{\pm0.019}$ & 0.65$^{\pm0.030}$ & 45.23$^{\pm2.14}$ & 0.58$^{\pm0.021}$ & 0.58$^{\pm0.018}$ & 0.64$^{\pm0.027}$ & 43.89$^{\pm1.96}$ \\

W/O IR  & 0.62$^{\pm0.023}$ & 0.62$^{\pm0.022}$ & 0.64$^{\pm0.026}$ & 44.18$^{\pm2.20}$ & 0.61$^{\pm0.022}$ & 0.60$^{\pm0.020}$ & 0.64$^{\pm0.028}$ & 44.35$^{\pm2.03}$ \\
W/O CoTR        & 0.60$^{\pm0.018}$ & 0.64$^{\pm0.022}$ & 0.65$^{\pm0.029}$ & 45.35$^{\pm2.15}$ & 0.61$^{\pm0.018}$ & 0.60$^{\pm0.021}$ & 0.64$^{\pm0.025}$ & 43.97$^{\pm1.79}$ \\
\textbf{Ours} & \textbf{0.69$^{\pm0.024}$} & \textbf{0.71$^{\pm0.027}$} & \textbf{0.75$^{\pm0.032}$} & \textbf{51.68$^{\pm2.36}$} & \textbf{0.68$^{\pm0.023}$} & \textbf{0.68$^{\pm0.022}$} & \textbf{0.72$^{\pm0.030}$} & \textbf{50.03$^{\pm2.14}$} \\

\bottomrule
\end{tabular}
\caption{
\textbf{Ablation study on HICO-DET of Open-Vocabulary annotations.}
}
\label{ablation}
\end{table*}

\begin{figure}
\centering 
\includegraphics[width=\linewidth]{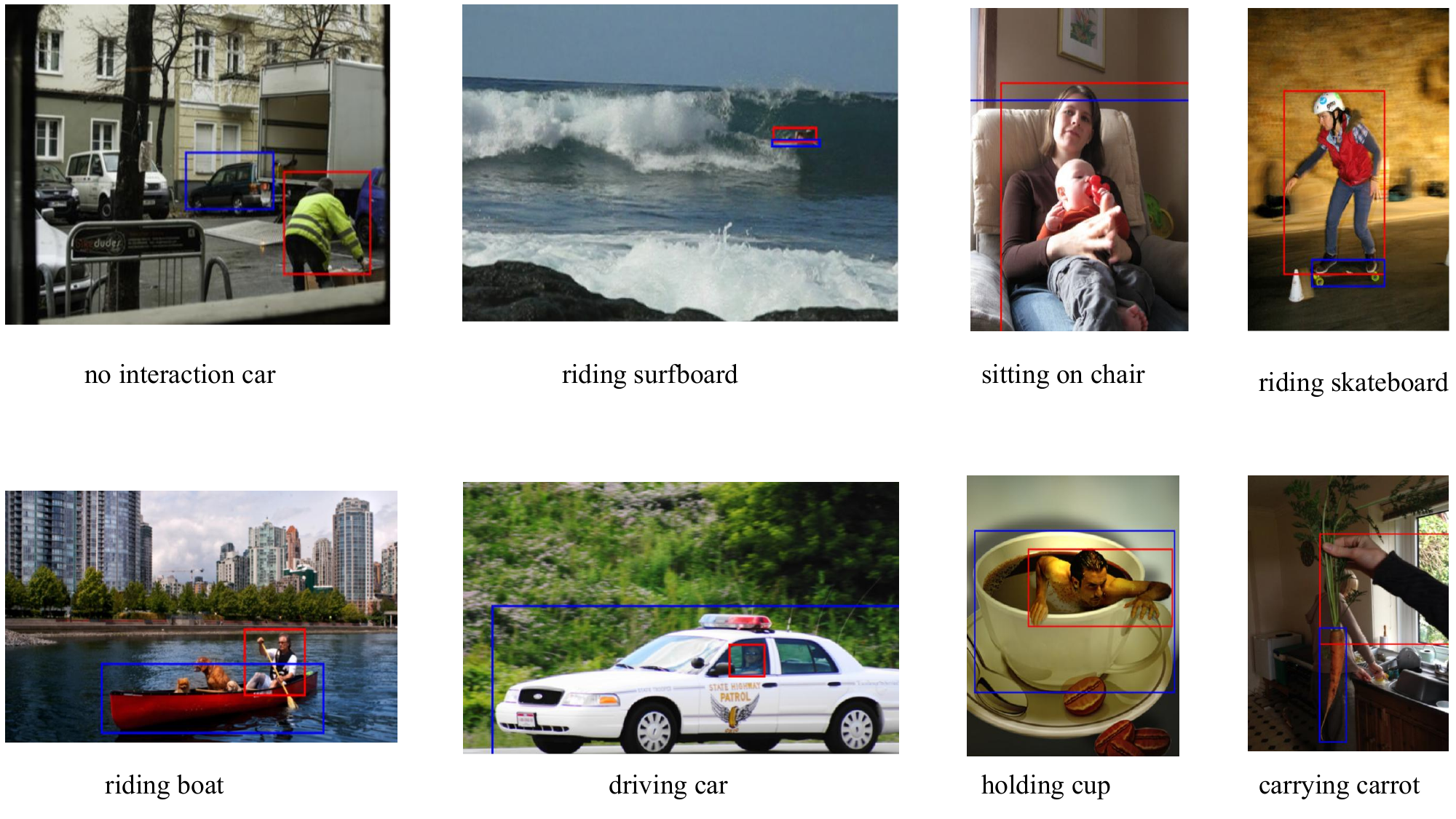}
\caption{\textbf{Qualitative result.}More Qualitative results in the Appendix. 
\label{vis}
}
\end{figure}
\subsection{Evaluation metric}
To accurately assess our model’s performance, we adopt four evaluation metrics distilled from prior work.

\begin{itemize}
    \item \textbf{H-mIOU} This metric is used to compute the mean Intersection over Union between the predicted person bounding boxes and the ground-truth.
    \item \textbf{O-mIOU} This metric calculates the mean Intersection over Union (mIoU) between the predicted object bounding boxes and their ground-truth.

    \item \textbf{A-ACC} This metric computes the probability of successfully predicting the interaction action.
    
    \item \textbf{mAP} A detection is deemed successful when both H-mIoU and O-mIoU exceed 0.5, and this metric measures the corresponding success rate.

\end{itemize}

Higher H-mIOU, O-mIOU, A-ACC and mAP mean powerful detection and reasoning ability of the method
\subsection{Main Result}
\textbf{Compare to SOTA method} We evaluate the four metrics on seen and unseen images from both datasets under all three annotation schemes, comparing our approach with recent state-of-the-art methods (e.g., HOI-Trans \cite{hoitrans}, DiffHOI\cite{diffhoi}, PAFR\cite{pafr}, HORP\cite{horp}). The results are summarized in Tables \ref{main-hico} and \ref{main-swig}.

Experimental results demonstrate that our model achieves state-of-the-art performance across all evaluation metrics. By leveraging the combined strengths of supervised fine-tuning and GRPO-based post-training, our approach exhibits robust open-world generalization, especially for combinations of open-vocabulary descriptions and unseen images.

\textbf{Qualitative results}
We present visualized results that demonstrate our model’s open-world generalization capabilities with Figure \ref{vis}, showing strong performance under open-vocabulary descriptions on unseen images. More Qualitative results in the Appendix.

\subsection{Ablation Study}
We perform ablation studies on the open-vocabulary annotations of the HICO-DET dataset to examine the contribution of each component to open-world HOI detection. The detailed results are presented in Table \ref{ablation}.

\textbf{W/O Post-training(W/O PT)}. We remove the GRPO-based post-training stage, which leads to a substantial drop in all four evaluation metrics. Without this reinforcement-learning fine-tuning, the model cannot leverage the multi-reward signals to refine its policy beyond the supervised fine-tuning stage, resulting in poorer localization (H-mIOU/O-mIOU) and degraded detection and action classification accuracy. 

% This confirms that post-training with GRPO is critical for harnessing the complementary reward functions and achieving robust open-world generalization.

\textbf{W/O Format Reward(W/O FR)}. We omit the format compliance reward, causing the model to generate HOI predictions with inconsistent or malformed ⟨subject, verb, object⟩ structures. Without this binary constraint, syntactic errors proliferate-missing tags, malformed tuples, and irregular delimiters-which in turn disrupt downstream parsing and degrade overall performance.

\textbf{W/O Detection Reward(W/O DR)}. We disable the detection reward (RIoU and Rreg), removing the incentive for precise bounding-box overlap and coordinate accuracy. Consequently, both H-mIOU and O-mIOU suffer significant declines, leading to lower mAP as the model neglects fine-grained localization in favor of other objectives. This shows that the detection reward is indispensable for driving accurate spatial predictions.

\textbf{W/O Interaction Reward(W/O IR)}. We turn off the interaction reward, so the model no longer receives feedback on correct action and object classification. As a result, A-ACC drops markedly, with the model frequently mislabeling verbs or objects despite reasonable bounding boxes. This indicates that the interaction reward is crucial for aligning the model’s predictions with the semantic ground truth.

\textbf{W/O CoT Reward(W/O CoTR)}. We remove the chain-of-thought reward, preventing optimization of the reasoning quality and coherence. The generated reasoning chains become semantically misaligned with the input prompt and logically disjointed across steps, leading to more hallucinations and less interpretable intermediate outputs. This validates that the CoT reward is vital for ensuring semantically meaningful and logically coherent reasoning trajectories.

\section{Conclusion}
This work presents HOID-R1, a unified framework for open-world human–object interaction detection that integrates chain-of-thought supervised fine-tuning with Group Relative Policy Optimization. By enforcing structured output formats and employing an MLLM-based judge to supervise intermediate reasoning, the approach mitigates hallucinations and grounds predictions in meaningful affordance cues. Extensive evaluations on HICO-DET and SWIG-HOI demonstrate that HOID-R1 outperforms existing methods in both seen and unseen settings under open-vocabulary evaluation, while ablation studies confirm the contribution of each component. Future research will focus on enhancing computational efficiency, extending the framework to video-based HOI detection for improved temporal coherence.

\appendix

\bibliography{aaai2026}

\end{document}